%% file: paper.tex
\documentclass{article}
\usepackage[final]{colm2026_conference}

\input{math_commands}

\usepackage[T1]{fontenc}
\usepackage{microtype}
\usepackage{hyperref}
\usepackage{url}
\usepackage{booktabs}
\usepackage{graphicx}
\usepackage{lineno}

\definecolor{darkblue}{rgb}{0, 0, 0.5}
\hypersetup{colorlinks=true, citecolor=darkblue, linkcolor=darkblue, urlcolor=darkblue}

\title{ConformalGuard: False-Positive-Controlled Action Gating for LM Classifiers}

\author{Bishnu Bista, Dhanraj Haibatrao Jadhav, \\
\textbf{Vibha Choudhary, Gaurav Saxena \& Arun Pandiyan Perumal} \\
Independent Researcher \\
\texttt{bishnubista@gmail.com, dhjadhav09@gmail.com} \\
\texttt{vibha.ch0511@gmail.com, gauravsaxena.oakland@gmail.com} \\
\texttt{arunpandiyanp@ieee.org}
\AND
Arjun Subedi \\
Astha.ai \\
\texttt{Arjun@astha.ai}
\AND
Jignesh Ammineni \\
University of Massachusetts Amherst \\
\texttt{jigneshammineni@gmail.com}}

\begin{document}

\ifcolmsubmission
\linenumbers
\fi

\maketitle

\begin{abstract}
Language model classifiers are increasingly deployed for high-stakes decisions such as phishing screening, content moderation, and fraud triage, yet validation-set thresholds provide no finite-sample control of deployment-time false positives. We present ConformalGuard, a model-agnostic calibration layer that wraps a fixed scoring function with finite-sample control over the rate at which legitimate inputs trigger an intervention tier. Our contribution is empirical and operational, not a new conformal theorem: zero-shot LM API scores can be coarse enough (10--11 unique values here) to make temperature scaling and isotonic regression operationally identical to raw thresholding for low-FPR action gating. We compare these methods with conformal calibration for fine-tuned BERT, Claude Sonnet 4, and GPT-4o on public email corpora. For Claude, conformal calibration reduces FPR from 0.040 to 0.008 while recall falls from 0.938 to 0.875. For deterministic GPT-4o scores, the 1\% operating point is too conservative to deploy (recall 0.564); three-run stochastic averaging raises recall to 0.732 at FPR 0.005. At 2\% prevalence with BERT, conformal calibration removes 25 false-positive reviews per 10,000 emails (95\% CI [10, 43]). The guarantee is legitimate-event-rate control under exchangeability; discrimination, workload effects, and robustness to shift remain empirical.
\end{abstract}

\section{Introduction}

Language models are increasingly used as classifiers for consequential decisions. An LLM screens emails as phishing or safe. A fine-tuned BERT flags toxic content. A GPT-based system triages fraud alerts. In each case, the deployment question is the same: given a score, when should the system act, defer, or escalate?

The standard approach is to tune a threshold on a validation set. This is a point estimate with no finite-sample coverage guarantee. It can overfit to the validation distribution, and the resulting false-positive rate at deployment may differ substantially from the target across scorer families.

The consequences at deployment are severe and asymmetric. Consider a security team processing 50,000 emails per day at 1\% phishing prevalence. A classifier tuned for 90\% recall with a 1\% false-positive rate sends 500 legitimate emails to analyst review daily. If the true rate is 1.5\%, a half-point miscalibration invisible in balanced evaluation, that count rises to 750. At realistic prevalence, the false-positive rate controls analyst workload, not recall. Model accuracy gains are eroded if the decision layer leaks false positives.

Standard post-hoc calibration methods, temperature scaling \citep{guo2017calibration} and isotonic regression \citep{zadrozny2002transforming}, improve probabilistic calibration on average but provide no finite-sample guarantee on the false-positive rate at a given threshold. Conformal prediction \citep{vovk2005algorithmic} offers exactly this: for any fixed scoring function and any target rate $\alpha$, it produces p-values satisfying Pr(p(X) $\leq$ $\alpha$ | Y=0) $\leq$ $\alpha$ under exchangeability, without distributional assumptions.

This paper asks a direct question: does conformal calibration provide better deployment-time false-positive control for LM classifiers than standard calibration methods when a threshold must be chosen before deployment labels are available?

We answer this by evaluating fine-tuned BERT, Claude Sonnet 4, and GPT-4o on the same public email corpus under four calibration strategies. Zero-shot LMs produce coarsely discretized scores: Claude returns only 11 unique values and deterministic GPT-4o returns 10. Temperature scaling and isotonic regression produce the same action boundary as raw validation-set tuning. At $\alpha=0.01$ on the in-distribution test split, the non-conformal alternatives miss the target by about 1.2 percentage points on average (mean absolute violation), with Claude reaching a 3.0-point overshoot, while conformal calibration stays within target. This does not imply improved discrimination or robustness under arbitrary shift; it provides a pre-deployment operating point under exchangeability.

Our contributions:

\begin{enumerate}
    \item \textbf{A systematic deployment comparison.} We evaluate raw thresholds, temperature scaling, isotonic regression, and conformal prediction across three LM scorers. In our in-distribution experiments, conformal calibration provides the tightest pre-deployment FPR control.

    \item \textbf{A score-granularity diagnosis.} Coarse zero-shot outputs make standard post-hoc calibration operationally inert for low-FPR action thresholds. Full-scale GPT-4o stochastic averaging shows that richer score grids improve the achievable recall--FPR trade-off.

    \item \textbf{Conformal action gating.} We map calibrated p-values to ordered intervention tiers (auto-clear, analyst review, specialist escalation, autonomous blocking) with cumulative legitimate-event-rate bounds for each tier or any stricter tier.

    \item \textbf{Operational impact at low prevalence.} With BERT at 2\% phishing prevalence, conformal calibration reduces false positives from 83 to 58 per 10,000 emails, a reduction of 25 reviews (95\% CI [10, 43]).
\end{enumerate}

\section{Background and related work}

\subsection{Post-hoc calibration of classifiers}

Temperature scaling \citep{guo2017calibration} learns a single scalar parameter T on a validation set, dividing logits by T before softmax. It improves expected calibration error on average but provides no guarantee that the FPR at any specific threshold is controlled. The learned temperature can overfit to the validation distribution.

Isotonic regression \citep{zadrozny2002transforming} fits a non-decreasing step function mapping scores to probabilities. It is non-parametric and handles arbitrary miscalibration shapes, but again provides no finite-sample FPR guarantee at a given operating point.

Both methods assume the input scores have sufficient granularity to support meaningful recalibration. When scores take only a handful of distinct values, as we find with zero-shot LLMs (Section 4), temperature scaling reduces to a monotone rescaling that cannot change the rank ordering and isotonic regression collapses to a lookup table with the same decision boundaries as the raw threshold.

LLM uncertainty estimation and confidence calibration span verbalized confidence, internal signals, consistency estimates, and ensemble methods \citep{xia2025survey,liu2025uncertainty}. Calibrating verbalized probabilities is especially relevant when an API exposes JSON probabilities rather than token log probabilities \citep{wang2024verbalized}. ConformalGuard is complementary: it does not make confidence intrinsically calibrated, but asks whether a frozen score supports a pre-deployment threshold with legitimate-event FPR control.

\subsection{Conformal prediction}

Conformal prediction \citep{vovk2005algorithmic, angelopoulos2023gentle} provides distribution-free, finite-sample guarantees under exchangeability. Let $C_0=\{x_i:Y_i=0\}_{i=1}^n$ contain only confirmed-legitimate examples, let $s_i=g(x_i)$ under a fixed scoring function $g$, and let $s(x)=g(x)$ for a new example. The conformal p-value is:

\begin{equation}
\label{eq:conformal-p}
p(x) = \frac{1 + \sum_{i=1}^{n} \mathbf{1}[s_i \geq s(x)]}{n + 1}
\end{equation}

Under exchangeability, this p-value provides a finite-sample bound on the false-positive rate at any chosen significance level $\alpha$ (stated formally as Proposition 1 in Section 3.2). This is not a probability of phishing. It is a rank-based conformity score that controls the rate at which legitimate examples fall below the threshold. The method does not require continuous scores or smooth calibration curves, but deterministic p-values inherit tied score blocks and therefore do not remove coarse-score plateaus.

Recent conformal work relevant to LMs includes ConU \citep{wang2024conu}, TECP \citep{xu2025tecp}, CADES \citep{zhang2025cades}, and the CP-for-NLP survey \citep{campos2024cpnlp}. \citet{bates2021distribution} and \citet{angelopoulos2024conformalrisk} extend conformal ideas to general risk control. \citet{tibshirani2019conformal} address covariate shift, while \citet{romano2020classification} develop adaptive conformal classification and randomized prediction sets.

Our contribution is not a new conformal theorem. It is an empirical deployment study of standard conformal and post-hoc calibration applied to LM classifier scores. Unlike randomized conformal classification, our deployed action rule is deterministic. Conformal calibration does not improve the score ordering; it selects a low-FPR operating point from legitimate calibration data before deployment labels are observed.

\subsection{Positioning}

\begin{table}[t]
\centering
\small
\setlength{\tabcolsep}{3pt}
\begin{tabular}{lccccc}
\toprule
Capability & Temp. scaling & Isotonic & ConU & CADES & \textbf{Ours} \\
\midrule
Finite-sample FPR guarantee & No & No & Yes & Yes & \textbf{Yes} \\
Validity on tied score grids & No & No & N/A & N/A & \textbf{Yes; utility limited} \\
Compared across LM architectures & No & No & No & No & \textbf{Yes} \\
Tiered action gating & No & No & No & No & \textbf{Cumulative bound} \\
\bottomrule
\end{tabular}
\caption{Positioning relative to existing work}
\label{tab:table1}
\end{table}

Table~\ref{tab:table1} summarizes our positioning. Our study isolates how coarse zero-shot LM score grids affect standard post-hoc calibration for low-FPR action gating across multiple scorer families.

\section{Method}

\subsection{Overview}

ConformalGuard wraps any fixed scoring function g(x) $\to$ [0,1] with a calibration layer that converts raw scores into calibrated p-values, then maps p-values to ordered action tiers.

\textbf{Step 1: Scoring.} A frozen scoring function g produces a suspiciousness score for each input. In our experiments, g is instantiated three ways: (a) fine-tuned BERT-base-uncased producing P(phishing), (b) Claude Sonnet 4 prompted to return a phishing probability, (c) GPT-4o prompted identically.

\textbf{Step 2: Calibration.} Given the legitimate-only set $C_0$ and scores $s_i=g(x_i)$ from the frozen scorer, we compute the conformal p-value in Equation~\ref{eq:conformal-p} for a test example $x$, with $s(x)=g(x)$.

For comparison, we also calibrate each scorer with: (a) a raw threshold tuned on the validation set, (b) temperature scaling learned on the validation set, and (c) isotonic regression fitted on the validation set.

\textbf{Step 3: Action gating.} Calibrated scores map to intervention tiers:

\begin{itemize}
    \item p > 0.10: auto-clear (no analyst involvement)
    \item 0.01 < p $\leq$ 0.10: analyst review
    \item 0.005 < p $\leq$ 0.01: specialist escalation
    \item p $\leq$ 0.005: reversible autonomous blocking
\end{itemize}

The cumulative legitimate-event rate for each tier or any stricter tier is bounded by the corresponding $\alpha$ threshold under exchangeability (Proposition 1).

\subsection{Conformal calibration}

\textbf{Proposition 1.} If the legitimate calibration and test examples are exchangeable, and the score function g is fixed before calibration, then Pr(p(X) $\leq$ $\alpha$ | Y=0) $\leq$ $\alpha$ for any $\alpha$ $\in$ [0,1].

\textit{Proof sketch.} Under exchangeability and in the absence of ties, the test score's rank among the n+1 scores (n calibration + 1 test) is uniform. With ties, the deterministic p-value based on the count of calibration scores greater than or equal to the test score assigns the least favorable rank within the tied block and is therefore conservative (super-uniform). Hence the probability that p is at most $\alpha$ is at most $\alpha$; see \citet[Chapter 2]{vovk2005algorithmic} and \citet[\S2]{angelopoulos2023gentle}.

We do not interpret p(x) as the probability that x is phishing. It is a rank-based conformity score relative to the legitimate calibration pool.

\textbf{What conformal adds on discretized scores.} Monotone post-hoc transformations cannot change the ranking or split a tied score block, so they often retain the validation-selected action boundary. Deterministic conformal p-values also do not create new discrimination or split ties. Their value is different: calibration ranks determine, before deployment, which existing score boundary satisfies the target legitimate-event rate under exchangeability. An oracle raw threshold chosen with test labels can match that boundary exactly. Randomized tie-breaking can split tied blocks, but sacrifices repeatable per-input tier assignments (Appendix~\ref{app:calibration-robustness}).

\textbf{Robustness to distribution shift.} If the total variation distance between calibration and test legitimate distributions is $\varepsilon$, then $\Pr(p(X)\leq\alpha\mid Y=0)\leq\alpha+\varepsilon$. Monitoring can detect a warning sign but does not restore validity under arbitrary shift. Persistent shift requires weighted, stratified, or adaptive conformal methods.

\textbf{Calibration set selection bias.} Confirmed-legitimate examples are not a random sample: they arise from auto-cleared events, analyst-reviewed events, or customer confirmations, each introducing selection bias. We mitigate through stratified sampling that matches the feature distribution between calibration and test sets, for example ensuring similar proportions of emails with full SMTP headers versus body-only text, since feature availability affects scorer behavior and exchangeability. Importance-weighted calibration \citep{tibshirani2019conformal} could further address this gap.

\textbf{Rolling calibration and model updates.} Each recalibration produces a new conformal procedure. After any scorer update, freeze the updated scorer, collect a fresh legitimate calibration set, rescore it, recompute ranks, and only then deploy. The ordering is train $\to$ validate $\to$ freeze $g$ $\to$ calibrate on disjoint legitimate data $\to$ deploy.

\subsection{What the guarantee does and does not cover}

\textbf{Controlled by design:} The rate at which legitimate events enter each action tier, under exchangeability.

\textbf{Not guaranteed:} Recall (detection rate), subgroup fairness, or workload reduction. These are supported empirically, not formally.

This is a deliberate scope restriction. A system that controls only legitimate-event rates is not a complete safety solution. But it addresses the deployment problem that causes the most operational pain: unnecessary analyst reviews of legitimate traffic.

\section{Experiments}

We evaluate four calibration strategies (raw threshold, temperature scaling, isotonic regression, and conformal prediction) across three scoring models (fine-tuned BERT, zero-shot Claude Sonnet 4, and zero-shot GPT-4o) on a unified public email phishing corpus.

\subsection{Data}

We assemble a public email corpus from four source families:

\begin{itemize}
    \item \textbf{Nazario Phishing Corpus} \citep{nazario2005phishing}. Verified real phishing with full SMTP headers.
    \item \textbf{Phishing Pot} \citep{phishingpot2022}. 7,069 raw honeypot phishing emails with Outlook/Exchange headers (4,393 in the final leakage-cleaned splits).
    \item \textbf{SpamAssassin} \citep{spamassassin2003corpus}. Public ham and spam corpus with full headers.
    \item \textbf{Enron-Spam AUEB} \citep{metsis2006spam}. Legitimate email baseline (body text only, no SMTP headers) and spam.
\end{itemize}

After excluding the EPVME synthetic dataset (identified as a protocol-vulnerability test suite containing XSS payloads rather than phishing emails; see supplementary audit report), filtering bodies below 50 characters, deduplicating by SHA-256 body hash, and removing 16 normalized-body duplicates from the lower-priority train/validation splits, the corpus contains 39,935 emails (Table~\ref{tab:splits}). Test and calibration were unchanged by the leakage cleanup.

\begin{table}[t]
\centering
\small
\begin{tabular}{lcccc}
\toprule
Split & Total & Legitimate & Phishing/Spam & Positive rate \\
\midrule
Train & 26,202 & 11,937 & 14,265 & 54.4\% \\
Val & 3,742 & 1,705 & 2,037 & 54.4\% \\
Cal & 2,000 & 2,000 & 0 & 0\% (legit. only) \\
Test & 7,991 & 3,911 & 4,080 & 51.1\% \\
\bottomrule
\end{tabular}
\caption{Corpus splits and class distribution (39,935 emails)}
\label{tab:splits}
\end{table}

The positive class includes both true phishing (Nazario, Phishing Pot; 28.7\% of positives) and spam (Enron spam, SpamAssassin spam; 71.3\%), following the convention of prior work on these corpora. The calibration set contains only confirmed-legitimate emails from multiple ham sources.

\textbf{Data quality.} A source-confusion classifier (TF-IDF + logistic regression, 5-fold CV) achieves 97.7\% accuracy at distinguishing sources from text. This is expected for public email corpora where phishing, spam, and legitimate emails have systematically different vocabulary. The calibration comparison is valid regardless because all four methods operate on the same score distributions. Cross-split duplicate check confirmed zero leakage after cleanup. The full audit report is in supplementary material.

\subsection{Scoring functions}

We evaluate three LM classifiers, each producing a phishing suspiciousness score s $\in$ [0,1]:

\begin{enumerate}
    \item \textbf{BERT-base-uncased (fine-tuned).} Fine-tuned on the training split for 3 epochs (lr=2e-5, batch=16, seed=42). Produces 1,176 unique score values across the corpus, a near-continuous distribution.

    \item \textbf{Claude Sonnet 4 (zero-shot).} Prompted with email subject and body to return a phishing probability. Produces \textbf{11 unique score values} (0.0, 0.05, 0.1, 0.2, 0.3, 0.7, 0.8, 0.85, 0.9, 0.95, 1.0), a coarsely discretized distribution.

    \item \textbf{GPT-4o (zero-shot).} Prompted identically. Produces \textbf{10 unique score values} (0.0, 0.1, 0.2, 0.3, 0.7, 0.8, 0.9, 0.95, 0.99, 1.0), similarly coarse.
\end{enumerate}

All three scoring functions are frozen before calibration. The sharp contrast in score granularity between BERT (1,176 levels) and the zero-shot LLMs (10--11 levels) is central to our results.

\subsection{Calibration methods}

For each scorer, we apply four calibration methods:

\begin{enumerate}
    \item \textbf{Raw threshold.} Choose threshold on the validation set to achieve target FPR $\leq$ $\alpha$. No post-hoc calibration.

    \item \textbf{Temperature scaling.} Learn T on the validation set by minimizing negative log-likelihood. Threshold the calibrated scores at the validation quantile matching $\alpha$.

    \item \textbf{Isotonic regression.} Fit a non-decreasing step function on the validation set. Threshold at the validation quantile matching $\alpha$.

    \item \textbf{Conformal prediction.} Compute conformal p-values on a disjoint calibration set of 2,000 legitimate-only examples. Threshold at $\alpha$.
\end{enumerate}

Methods 1 to 3 set their threshold on the validation set and hope it transfers to the test set. Method 4 sets its threshold on a separate calibration set with a finite-sample guarantee.

\subsection{Main result: calibration comparison across LMs}

\begin{table}[t]
\centering
\small
\begin{tabular}{lcccccc}
\toprule
LM & Calibration & Recall & Obs. FPR & FPR viol. & Precision & F1 \\
\midrule
BERT & Raw threshold & 0.996 $\pm$ 0.001 & 0.008 & $-0.002$ & 0.992 & 0.994 \\
BERT & Temp. scaling & 0.996 $\pm$ 0.001 & 0.008 & $-0.002$ & 0.992 & 0.994 \\
BERT & Isotonic reg. & 0.997 $\pm$ 0.001 & 0.013 & +0.003 & 0.988 & 0.993 \\
BERT & \textbf{Conformal} & \textbf{0.995 $\pm$ 0.001} & \textbf{0.006} & \textbf{$-0.004$} & \textbf{0.994} & \textbf{0.995} \\
Claude & Raw threshold & 0.938 $\pm$ 0.004 & 0.040 & +0.030 & 0.961 & 0.949 \\
Claude & Temp. scaling & 0.938 $\pm$ 0.004 & 0.040 & +0.030 & 0.961 & 0.949 \\
Claude & Isotonic reg. & 0.938 $\pm$ 0.004 & 0.040 & +0.030 & 0.961 & 0.949 \\
Claude & \textbf{Conformal} & \textbf{0.875 $\pm$ 0.005} & \textbf{0.008} & \textbf{$-0.002$} & \textbf{0.991} & \textbf{0.930} \\
GPT-4o & Raw threshold & 0.887 $\pm$ 0.005 & 0.015 & +0.005 & 0.984 & 0.933 \\
GPT-4o & Temp. scaling & 0.887 $\pm$ 0.005 & 0.015 & +0.005 & 0.984 & 0.933 \\
GPT-4o & Isotonic reg. & 0.887 $\pm$ 0.005 & 0.015 & +0.005 & 0.984 & 0.933 \\
GPT-4o & \textbf{Conformal} & \textbf{0.564 $\pm$ 0.007} & \textbf{0.001} & \textbf{$-0.009$} & \textbf{0.999} & \textbf{0.721} \\
\bottomrule
\end{tabular}
\caption{FPR control across three LMs and four calibration methods ($\alpha$ = 0.01)}
\label{tab:table2}
\end{table}

Table~\ref{tab:table2} reports 95\% bootstrap confidence intervals from 1,000 resamples. FPR violation = observed FPR minus target $\alpha$; negative means the method stays below the target, positive means it overshoots.

\textbf{Finding 1: conformal stays within target at $\alpha=0.01$ on the in-distribution split.} Across all three scorers, conformal calibration produces observed FPR below target. Every alternative overshoots on at least one scorer. The average absolute violation is 0.5pp for conformal versus about 1.2pp for the alternatives. Section~\ref{sec:shift} shows this does not imply validity under arbitrary source shift.

\textbf{Finding 2: Standard calibration methods are ineffective on LLM scores.} For Claude Sonnet 4, temperature scaling and isotonic regression produce results identical to raw thresholds: FPR 0.040 in all three cases (4x overshoot of the 0.01 target). The reason is mechanical. Claude returns only 11 distinct scores. Temperature scaling applies $s\to\operatorname{sigmoid}(\operatorname{logit}(s)/T)$, a monotone transform that preserves rank ordering. With only 11 input values, the transform maps to 11 output values, and the threshold falls between the same two adjacent scores regardless of $T$. Isotonic regression faces the same limitation. The three methods are operationally identical on discretized inputs.

\textbf{Finding 3: FPR control exposes a real utility trade-off.} Claude FPR falls from 0.040 to 0.008 while recall falls from 0.938 to 0.875. For deterministic GPT-4o, conformal reaches FPR 0.001 but recall falls from 0.887 to 0.564. This operating point is not deployable under a typical recall floor. Conformal selects a valid pre-deployment boundary; it does not improve classifier discrimination or overcome a poorly resolved score grid.

\textbf{Finding 4: BERT is the control case.} With 1,176 unique values, raw thresholding and temperature scaling behave similarly to conformal, while isotonic regression slightly overshoots (FPR 0.013). The small raw--conformal gap (0.008 vs. 0.006) supports score granularity, rather than the dataset alone, as the source of the larger LM gaps.

\subsection{Full-scale GPT-4o stochastic averaging}

The deterministic run is a fixed-scorer evaluation, but its ten-level output is unusually coarse. We therefore score the complete calibration, validation, and test splits at nonzero temperature and compare one stochastic draw with the mean of three independent draws, freezing each resulting pipeline before calibration (Table~\ref{tab:averaging}). Averaging increases unique scores from 33 to 105 and improves conformal recall from 0.285 to 0.732 at nearly unchanged FPR (0.0049 to 0.0051). The raw validation-selected threshold also improves, but still exceeds the 0.01 target (FPR 0.0194). Thus richer scoring improves the recall--FPR frontier; conformal calibration supplies threshold validity rather than the discrimination gain.

\begin{table}[t]
\centering
\small
\begin{tabular}{lrrrrrr}
\toprule
Runs & Unique & \multicolumn{3}{c}{Conformal} & \multicolumn{2}{c}{Raw threshold} \\
\cmidrule(lr){3-5}\cmidrule(lr){6-7}
averaged & scores & Recall & FPR & F1 & Recall & FPR \\
\midrule
1 & 33  & 0.285 & 0.0049 & 0.442 & 0.872 & 0.0355 \\
3 & 105 & 0.732 & 0.0051 & 0.843 & 0.882 & 0.0194 \\
\bottomrule
\end{tabular}
\caption{Full-scale GPT-4o stochastic averaging ($\alpha=0.01$)}
\label{tab:averaging}
\end{table}

\subsection{Low-prevalence evaluation}

At realistic inbox prevalence, false positives dominate analyst workload. We simulate 2\% phishing prevalence using BERT by keeping all 3,911 legitimate test emails and subsampling positives. We average prevalence-weighted precision over 50 repetitions; recall is measured on the full test positives, and false-positive intervals use paired bootstrap resampling of legitimate test emails (Table~\ref{tab:table3}).

\begin{table}[t]
\centering
\small
\begin{tabular}{lcccc}
\toprule
Calibration & Recall & FPR & Precision & FP per 10K emails (95\% CI) \\
\midrule
Raw threshold & 0.996 & 0.008 & 0.707 & 83 [55, 113] \\
Temp. scaling & 0.996 & 0.008 & 0.707 & 83 [55, 113] \\
\textbf{Conformal} & \textbf{0.995} & \textbf{0.006} & \textbf{0.776} & \textbf{58 [35, 83]} \\
\bottomrule
\end{tabular}
\caption{Low-prevalence evaluation (2\% phishing, BERT scorer)}
\label{tab:table3}
\end{table}

At 2\% prevalence, conformal calibration reduces false positives from 83 to 58 per 10,000 emails: 25 fewer reviews (paired-bootstrap 95\% CI [10, 43]). Precision rises from 0.707 to 0.776, while recall is statistically indistinguishable. Isotonic regression is omitted from this workload table because it already has the larger BERT FPR violation in Table~\ref{tab:table2}; including it would not affect the raw-versus-conformal workload comparison.

\subsection{Calibration set sensitivity}

How many legitimate examples are needed for effective conformal calibration? We subsample the calibration set for the BERT scorer at $\alpha$ = 0.01 (Table~\ref{tab:table4}).

\begin{table}[t]
\centering
\small
\begin{tabular}{lcc}
\toprule
Calibration n & FPR (mean $\pm$ std) & Recall (mean $\pm$ std) \\
\midrule
100 & 0.006 $\pm$ 0.006 & 0.988 $\pm$ 0.012 \\
500 & 0.006 $\pm$ 0.003 & 0.994 $\pm$ 0.002 \\
1,000 & 0.006 $\pm$ 0.002 & 0.995 $\pm$ 0.001 \\
2,000 (full) & 0.006 $\pm$ 0.000 & 0.995 $\pm$ 0.000 \\
\bottomrule
\end{tabular}
\caption{FPR and recall vs. calibration set size (BERT, $\alpha$ = 0.01)}
\label{tab:table4}
\end{table}

Mean FPR remains below target at $n=100$, but the high variance means individual calibration draws can exceed it. Recall stabilizes at $n\geq500$. We recommend at least 1,000 legitimate examples and full human review during cold start.

\section{Discussion and limitations}

\textbf{Calibration is not discrimination.} An oracle raw threshold selected with test labels can reproduce the deterministic conformal boundary because both act on the same score ordering. Conformal's advantage is pre-deployment selection using legitimate calibration data, not improved ranking. This distinction matters for GPT-4o: deterministic conformal recall drops from 0.887 to 0.564, so we would reject that scorer for 1\% FPR gating. Three-run averaging improves the frontier, but adds inference cost and requires the averaging procedure itself to be frozen before calibration.

\textbf{Randomization is a different interface.} Randomized conformal tie-breaking raises GPT-4o recall to 0.800 with mean FPR 0.011, a slight finite-test-split fluctuation above $\alpha=0.01$, but 17.2\% of test decisions depend on the random variate and seed-to-seed disagreement is 7.0\% (Appendix~\ref{app:calibration-robustness}). Deterministic tiers preserve repeatability for analyst-facing actions; randomized methods remain a useful population-level baseline rather than a free solution to coarse scores.

\subsection{Exchangeability and source shift}
\label{sec:shift}

The guarantee assumes exchangeable legitimate calibration and test examples. Table~\ref{tab:shift} isolates this limitation by calibrating and testing on different legitimate source families; within-source baselines appear in Appendix~\ref{app:calibration-robustness}. BERT calibrated on Enron ham and tested on SpamAssassin ham reaches FPR 0.021 at $\alpha=0.01$; GPT-4o reaches 0.017 in one reverse direction. Monitoring can trigger fallback or recalibration, but cannot restore validity under arbitrary shift. Persistent shift calls for stratified, weighted, or adaptive conformal methods \citep{tibshirani2019conformal,barber2023conformal}.

\begin{table}[t]
\centering
\small
\begin{tabular}{llllr}
\toprule
Scorer & Cal. source & Test source & $n_{\rm test}$ & FPR \\
\midrule
BERT & Enron & SpamAssassin & 801 & 0.021 \\
BERT & SpamAssassin & Enron & 3,110 & 0.003 \\
Claude & Enron & SpamAssassin & 801 & 0.006 \\
Claude & SpamAssassin & Enron & 3,110 & 0.008 \\
GPT-4o & Enron & SpamAssassin & 801 & 0.000 \\
GPT-4o & SpamAssassin & Enron & 3,110 & 0.017 \\
\bottomrule
\end{tabular}
\caption{All cross-source legitimate-shift directions ($\alpha=0.01$)}
\label{tab:shift}
\end{table}

\textbf{Dataset and model scope.} A source-confusion classifier reaches 97.7\% accuracy, so discrimination may partly exploit corpus artifacts; all calibration methods inherit that scorer. The positive class combines phishing (28.7\%) and spam (71.3\%), making the task malicious/unwanted-email detection rather than pure phishing. Claude and GPT-4o are zero-shot whereas BERT is fine-tuned, and our results do not establish how fine-tuned open LMs behave.

\textbf{What remains outside the guarantee.} We do not guarantee recall, workload reduction, adversarial robustness, or subgroup fairness. Attacks on positive inputs change detection rather than the legitimate-event FPR, while calibration-pool poisoning and legitimate-distribution attacks directly threaten exchangeability. Multiclass prediction sets, regression coverage, conditional fairness control, stronger attacks, and fine-tuned open-LM evaluations are future work.

\section{Conclusion}

We showed that conformal calibration can select tighter pre-deployment false-positive operating points than validation-selected temperature scaling and isotonic regression for LM classifiers with coarse score grids. This is a standard conformal guarantee used as an operational action-gating interface, not a new theorem or a claim of improved discrimination.

For Claude Sonnet 4, FPR falls from 0.040 to 0.008 at a 6.3-point recall cost. Deterministic GPT-4o exposes an unacceptable recall loss, while three-run averaging recovers recall to 0.732 at FPR 0.005. At 2\% prevalence with BERT, calibration avoids 25 false-positive reviews per 10,000 emails (95\% CI [10, 43]). These are operating-point choices whose validity depends on exchangeability and whose utility must be evaluated separately.

The practical rule is simple: freeze the complete scorer, calibrate on fresh legitimate examples, gate actions by p-value tier, and reject the scorer when the target FPR implies unacceptable recall. After any model or prompting change, rescore and recalibrate before deployment.

\section*{Reproducibility statement}

The evaluation uses exclusively public datasets: Nazario Phishing Corpus \citep{nazario2005phishing}, Phishing Pot \citep{phishingpot2022}, SpamAssassin \citep{spamassassin2003corpus}, and Enron-Spam AUEB \citep{metsis2006spam}. The EPVME dataset was excluded after a quality audit identified it as a protocol-vulnerability test suite rather than a phishing corpus; the audit report is in supplementary material. The public camera-ready artifact is available at \url{https://github.com/bishnubista/conformalguard-colm-2026}. The package contains pinned data-download and parsing scripts, SHA-256 provenance, the leakage-cleaned split manifest, exact training and prompting configurations, calibration code, score-only records, and reported outputs. The public export omits raw messages and parsed email bodies.

\section*{LLM use disclosure}

LLMs are used in two capacities: (1) as classification systems under study (fine-tuned BERT, Claude Sonnet 4, GPT-4o) and (2) for writing assistance. All reported numbers are verified against bundled output files. The evaluation prompts are included in the repository.

\section*{Ethics statement}

All evaluation data comes from publicly released research corpora: Nazario Phishing Corpus \citep{nazario2005phishing}, Phishing Pot \citep{phishingpot2022}, SpamAssassin \citep{spamassassin2003corpus}, and Enron-Spam AUEB \citep{metsis2006spam}. No customer data was used. Because these corpora can contain real messages and personal information, the public artifact omits raw messages and parsed bodies and instead provides pinned retrieval scripts, hashes, pseudonymous score records, and aggregate outputs. The EPVME synthetic dataset was excluded on data-quality grounds. Production deployment would require data minimization, regulatory compliance, and fairness auditing for disparate impact.

\bibliography{references}
\bibliographystyle{colm2026_conference}

\appendix
\section{Calibration sensitivity and robustness analyses}
\label{app:calibration-robustness}

\subsection{Class-balanced calibration ablation}
Adding positives when fitting temperature scaling and isotonic regression does not change the Claude or GPT-4o decisions at $\alpha=0.01$ (Table~\ref{tab:balanced}). Score granularity, not class composition of the fit set, remains the bottleneck.

\begin{table}[b]
\centering
\small
\begin{tabular}{llrrrr}
\toprule
Scorer & Method & Recall & FPR & Precision & F1 \\
\midrule
BERT & Temp. & 0.9963 & 0.0084 & 0.9919 & 0.9941 \\
BERT & Isotonic & 0.9973 & 0.0130 & 0.9876 & 0.9924 \\
Claude & Temp. & 0.9377 & 0.0399 & 0.9608 & 0.9491 \\
Claude & Isotonic & 0.9377 & 0.0399 & 0.9608 & 0.9491 \\
GPT-4o & Temp. & 0.8870 & 0.0153 & 0.9837 & 0.9329 \\
GPT-4o & Isotonic & 0.8870 & 0.0153 & 0.9837 & 0.9329 \\
\bottomrule
\end{tabular}
\caption{Balanced-validation calibration ablation}
\label{tab:balanced}
\end{table}

\subsection{Sensitivity to the target false-positive rate}
Table~\ref{tab:alpha} shows smooth movement for BERT but plateaus for coarse LM scores. Claude is unchanged between $\alpha=0.01$ and 0.05; GPT-4o is unchanged from 0.001 through 0.01.

\begin{table}[t]
\centering
\small
\begin{tabular}{lrrrrrrrrrr}
\toprule
& \multicolumn{2}{c}{$\alpha=.001$} & \multicolumn{2}{c}{$.005$} & \multicolumn{2}{c}{$.01$} & \multicolumn{2}{c}{$.05$} & \multicolumn{2}{c}{$.10$} \\
\cmidrule(lr){2-3}\cmidrule(lr){4-5}\cmidrule(lr){6-7}\cmidrule(lr){8-9}\cmidrule(lr){10-11}
Scorer & Rec. & FPR & Rec. & FPR & Rec. & FPR & Rec. & FPR & Rec. & FPR \\
\midrule
BERT & .968 & .001 & .992 & .003 & .995 & .006 & .999 & .052 & 1.000 & .097 \\
Claude & .040 & .000 & .729 & .002 & .875 & .008 & .875 & .008 & .970 & .060 \\
GPT-4o & .564 & .001 & .564 & .001 & .564 & .001 & .887 & .015 & .953 & .067 \\
\bottomrule
\end{tabular}
\caption{Deterministic conformal threshold sensitivity}
\label{tab:alpha}
\end{table}

\subsection{Within-source false-positive baselines}
For Enron ($n=3{,}110$) and SpamAssassin ($n=801$), respectively, the matched-source FPRs are BERT 0.006/0.011, Claude 0.008/0.006, and GPT-4o 0.001/0.008. The BERT SpamAssassin result is a one-count-scale finite-test-split overshoot at $\alpha=0.01$, not evidence that matched source alone guarantees exact empirical control.

\subsection{Randomized conformal tie-breaking}
For $G$ calibration scores greater than a test score and $E$ equal to it, we use $p=(G+U(E+1))/(n+1)$ with $U\sim\mathrm{Unif}(0,1)$ and report 100-seed means (Table~\ref{tab:randomized}). Randomization improves GPT-4o recall but changes the per-input decision contract; its mean test FPR of 0.011 is a finite-test-split fluctuation around $\alpha=0.01$.

\begin{table}[ht]
\centering
\small
\begin{tabular}{lrrrrr}
\toprule
Scorer & Recall & FPR & F1 & Depends on $U$ & Seed disagreement \\
\midrule
BERT & 0.995 & 0.006 & 0.995 & 0.0\% & 0.0\% \\
Claude & 0.881 & 0.010 & 0.932 & 4.7\% & 0.7\% \\
GPT-4o & 0.800 & 0.011 & 0.883 & 17.2\% & 7.0\% \\
\bottomrule
\end{tabular}
\caption{Randomized conformal tie-breaking at $\alpha=0.01$}
\label{tab:randomized}
\end{table}

\end{document}

%% file: math_commands.tex
\usepackage{amsmath,amsfonts,bm}

\def\eqref#1{equation~\ref{#1}}

\def\1{\bm{1}}

\DeclareMathAlphabet{\mathsfit}{\encodingdefault}{\sfdefault}{m}{sl}
\SetMathAlphabet{\mathsfit}{bold}{\encodingdefault}{\sfdefault}{bx}{n}













%% file: paper.bbl
\begin{thebibliography}{20}
\providecommand{\natexlab}[1]{#1}
\providecommand{\url}[1]{\texttt{#1}}
\expandafter\ifx\csname urlstyle\endcsname\relax
  \providecommand{\doi}[1]{doi: #1}\else
  \providecommand{\doi}{doi: \begingroup \urlstyle{rm}\Url}\fi

\bibitem[Angelopoulos \& Bates(2023)Angelopoulos and
  Bates]{angelopoulos2023gentle}
Anastasios~N. Angelopoulos and Stephen Bates.
\newblock A gentle introduction to conformal prediction and distribution-free
  uncertainty quantification.
\newblock \emph{Foundations and Trends in Machine Learning}, 16\penalty0
  (4):\penalty0 494--591, 2023.
\newblock \doi{10.1561/2200000101}.

\bibitem[Angelopoulos et~al.(2024)Angelopoulos, Bates, Fisch, Lei, and
  Schuster]{angelopoulos2024conformalrisk}
Anastasios~N. Angelopoulos, Stephen Bates, Adam Fisch, Lihua Lei, and Tal
  Schuster.
\newblock Conformal risk control.
\newblock In \emph{International Conference on Learning Representations
  (ICLR)}, 2024.

\bibitem[{Apache SpamAssassin Project}(2003)]{spamassassin2003corpus}
{Apache SpamAssassin Project}.
\newblock The {SpamAssassin} public mail corpus.
\newblock \url{https://spamassassin.apache.org/old/publiccorpus/}, 2003.

\bibitem[Barber et~al.(2023)Barber, Candes, Ramdas, and
  Tibshirani]{barber2023conformal}
Rina~Foygel Barber, Emmanuel~J. Candes, Aaditya Ramdas, and Ryan~J. Tibshirani.
\newblock Conformal prediction beyond exchangeability.
\newblock \emph{Annals of Statistics}, 51\penalty0 (2):\penalty0 816--845,
  2023.
\newblock \doi{10.1214/23-AOS2276}.

\bibitem[Bates et~al.(2021)Bates, Angelopoulos, Lei, Malik, and
  Jordan]{bates2021distribution}
Stephen Bates, Anastasios Angelopoulos, Lihua Lei, Jitendra Malik, and
  Michael~I. Jordan.
\newblock Distribution-free, risk-controlling prediction sets.
\newblock \emph{Journal of the ACM}, 68\penalty0 (6):\penalty0 1--34, 2021.
\newblock \doi{10.1145/3478535}.

\bibitem[Campos et~al.(2024)Campos, Farinhas, Zerva, Figueiredo, and
  Martins]{campos2024cpnlp}
Margarida~M. Campos, Ant{\'o}nio Farinhas, Chrysoula Zerva, M{\'a}rio A.~T.
  Figueiredo, and Andr{\'e} F.~T. Martins.
\newblock Conformal prediction for natural language processing: A survey.
\newblock \emph{Transactions of the Association for Computational Linguistics},
  12:\penalty0 1497--1516, 2024.
\newblock \doi{10.1162/tacl_a_00715}.

\bibitem[Guo et~al.(2017)Guo, Pleiss, Sun, and Weinberger]{guo2017calibration}
Chuan Guo, Geoff Pleiss, Yu~Sun, and Kilian~Q. Weinberger.
\newblock On calibration of modern neural networks.
\newblock In \emph{Proceedings of the 34th International Conference on Machine
  Learning (ICML)}, volume~70, pp.\  1321--1330. PMLR, 2017.

\bibitem[Liu et~al.(2025)Liu, Chen, Da, Chen, Lin, and Wei]{liu2025uncertainty}
Xiaoou Liu, Tiejin Chen, Longchao Da, Chacha Chen, Zhen Lin, and Hua Wei.
\newblock Uncertainty quantification and confidence calibration in large
  language models: A survey.
\newblock \emph{arXiv preprint arXiv:2503.15850}, 2025.

\bibitem[Metsis et~al.(2006)Metsis, Androutsopoulos, and
  Paliouras]{metsis2006spam}
Vangelis Metsis, Ion Androutsopoulos, and Georgios Paliouras.
\newblock Spam filtering with naive {Bayes} -- which naive {Bayes}?
\newblock In \emph{CEAS 2006 -- Third Conference on Email and Anti-Spam}, 2006.

\bibitem[Nazario(2005)]{nazario2005phishing}
Jos{\'e} Nazario.
\newblock Nazario phishing corpus.
\newblock \url{https://monkey.org/~jose/phishing/}, 2005.
\newblock Public phishing mailbox collection.

\bibitem[{rf-peixoto}(2022)]{phishingpot2022}
{rf-peixoto}.
\newblock Phishing pot.
\newblock \url{https://github.com/rf-peixoto/phishing_pot}, 2022.
\newblock Real-world honeypot phishing email collection.

\bibitem[Romano et~al.(2020)Romano, Sesia, and
  Candes]{romano2020classification}
Yaniv Romano, Matteo Sesia, and Emmanuel Candes.
\newblock Classification with valid and adaptive coverage.
\newblock In \emph{Advances in Neural Information Processing Systems
  (NeurIPS)}, volume~33, 2020.

\bibitem[Tibshirani et~al.(2019)Tibshirani, Barber, Candes, and
  Ramdas]{tibshirani2019conformal}
Ryan~J. Tibshirani, Rina~Foygel Barber, Emmanuel~J. Candes, and Aaditya Ramdas.
\newblock Conformal prediction under covariate shift.
\newblock In \emph{Advances in Neural Information Processing Systems
  (NeurIPS)}, volume~32, 2019.

\bibitem[Vovk et~al.(2005)Vovk, Gammerman, and Shafer]{vovk2005algorithmic}
Vladimir Vovk, Alex Gammerman, and Glenn Shafer.
\newblock \emph{Algorithmic Learning in a Random World}.
\newblock Springer, New York, 2005.
\newblock \doi{10.1007/b106715}.

\bibitem[Wang et~al.(2024{\natexlab{a}})Wang, Szarvas, Balazs, Danchenko, and
  Ernst]{wang2024verbalized}
Cheng Wang, Gyuri Szarvas, Georges Balazs, Pavel Danchenko, and Patrick Ernst.
\newblock Calibrating verbalized probabilities for large language models.
\newblock \emph{arXiv preprint arXiv:2410.06707}, 2024{\natexlab{a}}.

\bibitem[Wang et~al.(2024{\natexlab{b}})Wang, Duan, Cheng, Zhang, Wang, Shi,
  Xu, Shen, and Zhu]{wang2024conu}
Zhiyuan Wang, Jinhao Duan, Lu~Cheng, Yue Zhang, Qingni Wang, Xiaoshuang Shi,
  Kaidi Xu, Heng~Tao Shen, and Xiaofeng Zhu.
\newblock {ConU}: Conformal uncertainty in large language models with
  correctness coverage guarantees.
\newblock In \emph{Findings of the Association for Computational Linguistics:
  EMNLP 2024}, pp.\  6886--6898. Association for Computational Linguistics,
  2024{\natexlab{b}}.
\newblock \doi{10.18653/v1/2024.findings-emnlp.404}.

\bibitem[Xia et~al.(2025)Xia, Xu, Zhang, and Liu]{xia2025survey}
Zhiqiu Xia, Jinxuan Xu, Yuqian Zhang, and Hang Liu.
\newblock A survey of uncertainty estimation methods on large language models.
\newblock \emph{arXiv preprint arXiv:2503.00172}, 2025.

\bibitem[Xu \& Lu(2025)Xu and Lu]{xu2025tecp}
Beining Xu and Yongming Lu.
\newblock {TECP}: Token-entropy conformal prediction for {LLMs}.
\newblock \emph{arXiv preprint arXiv:2509.00461}, 2025.
\newblock URL \url{https://arxiv.org/abs/2509.00461}.

\bibitem[Zadrozny \& Elkan(2002)Zadrozny and Elkan]{zadrozny2002transforming}
Bianca Zadrozny and Charles Elkan.
\newblock Transforming classifier scores into accurate multiclass probability
  estimates.
\newblock In \emph{Proceedings of the Eighth ACM SIGKDD International
  Conference on Knowledge Discovery and Data Mining}, pp.\  694--699. ACM,
  2002.
\newblock \doi{10.1145/775047.775151}.

\bibitem[Zhang et~al.(2025)Zhang, Zhou, Liu, Zhang, Lin, and
  Pan]{zhang2025cades}
Shuai Zhang, Chuan Zhou, Yang Liu, Peng Zhang, Xixun Lin, and Shirui Pan.
\newblock Conformal anomaly detection in event sequences.
\newblock In \emph{Proceedings of the 42nd International Conference on Machine
  Learning}, volume 267 of \emph{Proceedings of Machine Learning Research},
  pp.\  77177--77193. PMLR, 2025.

\end{thebibliography}
